%% file: conference_101719.tex
\documentclass[conference]{IEEEtran}
\IEEEoverridecommandlockouts

\usepackage{etoolbox}
\makeatletter
\patchcmd{\@makecaption}{\scshape}{}{}{}
\makeatother
\usepackage{mathtools}
\usepackage{amssymb}
\usepackage{pifont}
\newcommand{\cmark}{\ding{51}}%
\newcommand{\xmark}{\ding{55}}%
\usepackage{hyperref}
\usepackage{cite}
\usepackage{amsmath,amssymb,amsfonts}
\usepackage{algorithmic}
\usepackage{graphicx}
\usepackage{import}
\usepackage{textcomp}
\usepackage{multirow}
\usepackage{xcolor,colortbl}
\usepackage{cleveref}

\usepackage{booktabs}
\usepackage{enumitem}
\definecolor{inchworm}{rgb}{0.7, 0.93, 0.36}
\definecolor{gold}{rgb}{1.0, 0.84, 0.0}
\definecolor{lemon}{rgb}{1.0, 0.97, 0.0}
\usepackage{subcaption}
\usepackage{contour}
\contourlength{0.1pt} 
\usepackage{comment}
   
\newcommand{\fs}{ \cellcolor{gold}\bf }   
\newcommand{\nd}{ \cellcolor{lemon}\underline }      

\newcommand{\PAR}[1]{\vskip0pt \noindent{\bf #1~}}
\def\BibTeX{{\rm B\kern-.05em{\sc i\kern-.025em b}\kern-.08em
    T\kern-.1667em\lower.7ex\hbox{E}\kern-.125emX}}
\begin{document}
\setlength{\skip\footins}{0.1cm}
\title{SpotLight:  Robotic Scene Understanding through Interaction and Affordance Detection
}

\author{\IEEEauthorblockN{Tim Engelbracht$^{1}$, René Zurbrügg$^{1}$, Marc Pollefeys$^{1,2}$, Hermann Blum$^{1,3,\dagger}$, Zuria Bauer$^{1,\dagger}$}

\thanks{$^1$ETH Zürich, $^2$Microsoft, $^3$ Uni Bonn, $\dagger$ denotes equal advising.}
\thanks{Corresponding Author: Tim Engelbracht $<$\href{mailto:tengelbracht@ethz.ch}{tengelbracht@ethz.ch}$>$} \thanks{This research is partially supported by the ETH AI Center and by the Lamarr Institute for Machine Learning and Artificial Intelligence.}
}

\maketitle
\begin{abstract}
Despite increasing research efforts on household robotics, robots intended for deployment in domestic settings still struggle with more complex tasks such as interacting with functional elements like drawers or light switches, largely due to limited task-specific understanding and interaction capabilities.
These tasks require not only detection and pose estimation but also an understanding of the affordances these elements provide. To address these challenges and enhance robotic scene understanding, we introduce SpotLight: A comprehensive framework for robotic interaction with functional elements, specifically light switches. Furthermore, this framework enables robots to improve their environmental understanding through interaction. Leveraging VLM-based affordance prediction to estimate motion primitives for light switch interaction, we achieve up to 84\% operation success in real world experiments. We further introduce a specialized dataset containing 715 images as well as a custom detection model for light switch detection. We demonstrate how the framework can facilitate robot learning through physical interaction by having the
robot explore the environment and discover previously unknown relationships in a scene graph representation. Lastly, we propose an extension to the framework to accommodate other functional interactions such as swing doors, showcasing its flexibility.\\
Videos and Code: \url{timengelbracht.github.io/SpotLight/}


\end{abstract}


\section{Introduction} 
Despite robots making their way into our living spaces as cleaning robots (e.g. Roomba) and lab demos of pick-and-place assistants~\cite{wu2023tidybot, lemke2024spot}, fully capable domestic assistants are still far out of reach. One key capability missing is interaction with the environment, in particular through functional elements like thermostats, door knobs or light switches.

To interact with functional elements, robots not only need to detect the type of object they interact with, but also need to understand what kind of physical interaction this element allows for - often called its \emph{affordance} \cite{Gibson1979-GIBTEA}. Amplifying this difficulty, functional elements are usually small, making them hard to detect directly from point clouds even if cm resolution is available. Aside from perception, the interaction requires fine-grained and physically compliant manipulation by the robot. Finally, to function autonomously in domestic environments, robots need to have an understanding of the effect a functional element has on its environment (e.g.  what happens to the drawer if the handle is pulled, what lamp turns on if the light switch is operated). While mobile manipulation has received a lot of attention in recent years, most of the work is focused on planning and control \cite{Xia_2020, Kindle_2020} or uses off-the-shelf grasping algorithms for Go-Fetch tasks \cite{lemke2024spot, Blomqvist_2020}. Although impressive in their respective domains, they fall short of equipping the mobile manipulator with the affordance capabilities needed for fine-grained interaction with functional elements.
\begin{figure}[t!] 
    \centering
    \includegraphics[width=1.0\linewidth]{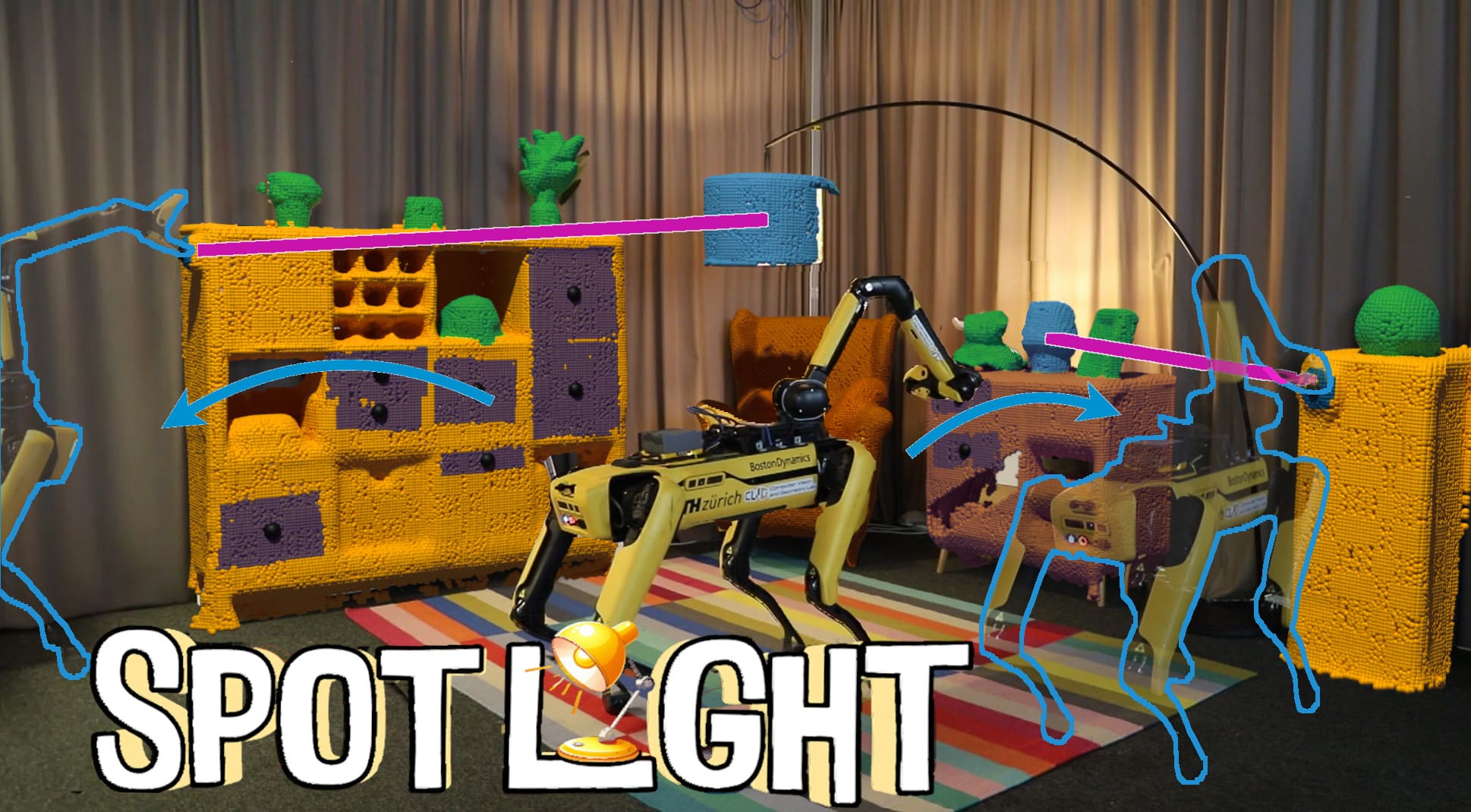}
    \caption{\textbf{SpotLight}. By interacting with different functional elements such as light-switches in the scene and observing the induced change, the robot is able to learn the correct relations between the lamps and light-switches in the scene. 
    }
    \label{fig:enter-label}
\end{figure}

In this paper we introduce \emph{SpotLight}: A framework for robotic interaction with light switches that enables a robot to enhance its understanding of the environment through interaction. Our method utilizes a mobile agent's RGB-D camera to detect light switches in images with a detector trained on a newly collected light switch dataset. Once detected, we then propose a method to accurately register the 3D pose of the element in the environment.
Afterwards, it leverages a Vision-Language-Model (VLM) for affordance prediction and enables fine-grained interaction by fusing 3D pose and 2D affordance prediction. We further demonstrate how the framework can facilitate robotic learning through interaction by having the robot explore and discover previously unknown relationships between lamps and light switches.
Our main contributions are as follows:

\begin{itemize}
    \item A framework for mobile interaction with functional elements, specifically light switches, as well as extensive real-world evaluations. 
    \item An annotated light-switch data set containing 715 images and  Yolov8 fine-tuned for light-switch detection.
    \item An application to embodied perception by learning light-switch and lamp connections in a scene graph.
\end{itemize}

\newpage
\section{Related Work}\vspace{-0.1cm}
This work focuses on inferring and describing motions for interaction with functional elements, closely related to affordance prediction and 3D motion estimation. Additionally, through physical interaction, we aim for a mobile agent to learn hidden environmental relationships, commonly referred as interactive perception. 
\PAR{Affordance Prediction and Motion Estimation.} The term \textit{affordance}\cite{Gibson1979-GIBTEA} 
describes the potential physical interactions that an object or part of it enables for an agent (e.g. a chair \textit{affords} sitting, a button pushing). 
Building on this concept, the field of affordance prediction deals with inferring affordances of objects from images and video \cite{Fang_2018_CVPR, Do_2017, Luddecke_2017_ICCV, Nagarajan_2018} or 3D representations \cite{delitzas2024scenefun3d, xu2022, Nagarajan_2020, mo2021, Deng_2021}. 
Affordance prediction has sparked interest in the robotics community as a prior for tasks such as robotic grasping \cite{mandikal_2020, Kokic_2017, Redmon_2014} in pick-and-place settings \cite{Wu_2020, Lin_2021} and even planning/exploration \cite{Nagarajan_2020, Xu_2020}. However, while works like \cite{mandikal_2020} show impressive results in dexterous grasping by predicting affordance regions, they offer no further interaction description (i.e. what to do with the object once it is grasped) which is key for robotic interaction with functional elements. Recent works \cite{delitzas2024scenefun3d} offer more fine-grained interaction descriptions and motion estimates for functional elements but require high fidelity (laser-scanned) point clouds to capture detailed functional elements in 3D, thus limiting deployment significantly. Infering fine-grained affordances from commodity RGB-D point clouds \cite{Dai_2017, chang_2017} is challenging due to the insufficient resolution \cite{delitzas2024scenefun3d}. Our method directly predicts affordances from the high resolution camera image of a mobile agent using a Vision-Language Model (VLM) and combining them with heuristics for 3D motion estimation. Leveraging the VLM's semantic understanding for affordance prediction allows us to bypass the need for high-fidelity pointclouds.

The related field of motion estimation explores the actual 3D motion for an object interaction (e.g. estimating the motion of a cabinet drawer). One line of works focuses on articulated objects and estimating their motion parameters \cite{Li_2019, Hu_2017, jiang2022, Yan_2020, Wang_2019} from either Depth, 3D or RGB-D. Focusing mainly on articulated motions, works like \cite{sun2023, jiang2022} deal with interactable elements such as handles, but focus on openable parts, thus neglecting other functional elements like light switches.

\PAR{Interactive Perception.}
Interactive perception refers to the task of an agent interacting with its environment and thereby gaining new insights of the environment \cite{Bohg_2016}. Most prominently, it has been used to learn object articulations \cite{nie2023structureactionlearninginteractions, hsu2023, jiang2022z, katz_2023, Hausman_2015, Martimartin_2014}, but has also been applied to other tasks such as mass distribution recovery through interaction \cite{kumar2020}. In this work, we extend interactive perception to learning the associations between light switches and lamps — relations that are inherently unobservable without interaction. Although such interactions do not cause immediate object deformation or measurable motion, they result in detectable environmental changes, such as variations in light intensity. By monitoring the environment before and after interaction, the agent learns new relationships in the scene and dynamically updates a scene graph representation \cite{Chang_2023, rosinol_2020}.

\vspace{-0.2cm}
\section{Method} \vspace{-0.1cm}
\PAR{Problem Formulation.}
We consider a mobile agent equipped with a robotic arm and a two-finger gripper. 
We further assume that a reconstruction of the environment from SLAM or camera scans is available\footnote{We use scans from a consumer-grade rgbd camera (i.e. Ipad)}. This reconstruction is represented as a pointcloud $P_{env} = \{(\vec{p}_i, c_i)\}_{i=0}^{N_{pts}}$ where each point $\vec{p}_i \in \mathbb{R}^3$ has a corresponding color and an object-level semantic label $c_i \in \{ {1,\dots,N_c}$ \}. 
The question we aim to answer in this work is:
i) How can we interact with different functional objects in this environment.
More specifically, we consider the interaction with various kinds of \emph{light switches} and \emph{swing doors}. ii) How can we leverage object interactions to learn about unobservable states and functional relations, such as the relationship between light switches and lamps.

\begin{figure*}
\vspace{-0.9cm}
    \centering
    \includegraphics[width=\linewidth]{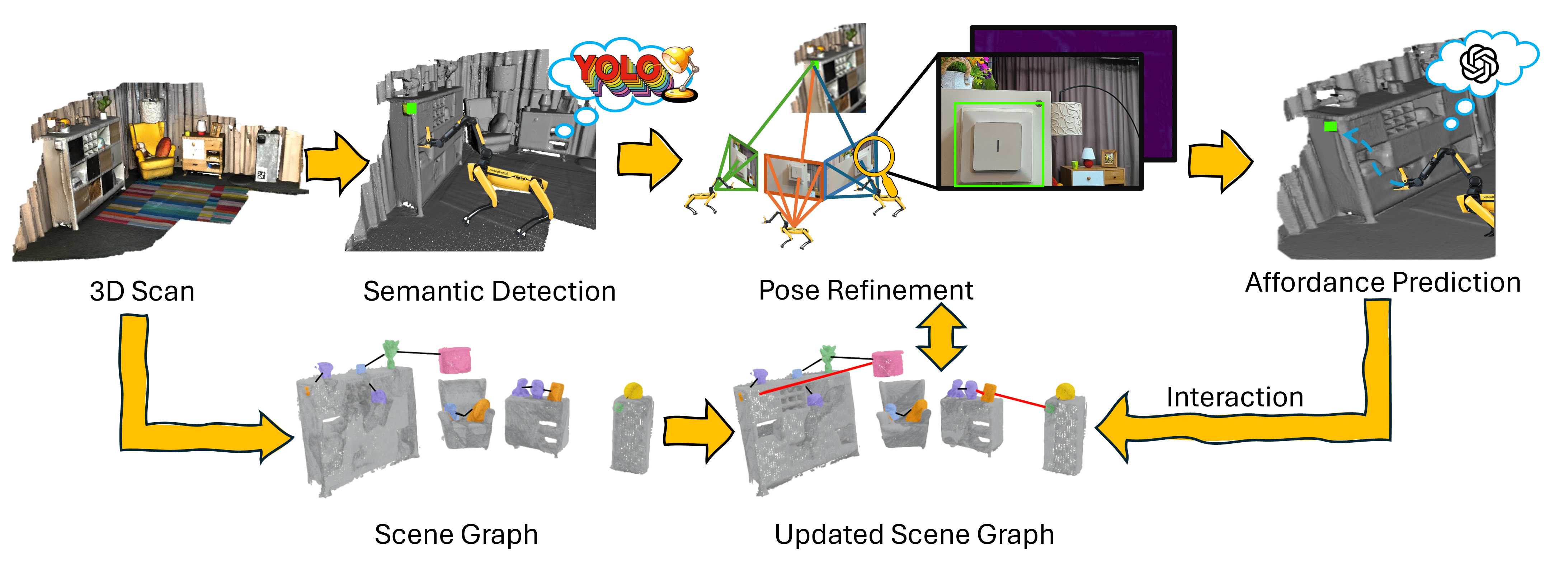}
    \vspace{-0.6cm}
    \caption{\textbf{Method Overview.} Given a point cloud reconstruction of the environment, we simultaneously perform instance segmentation and extract a scene graph representation. We then detect functional elements (e.g. light switches) in 2D, and further register and refine their pose in 3D. Thereafter we predict their functional affordance (e.g. $\texttt{"rotate X$^{\circ}$"}$) and estimate 3D motion primitives for interaction. Given initially hidden relationships between functional elements and objects in the scene (e.g. which light switch operates which lamp), the robot explores these through interaction with the environment.
    \vspace{-0.6cm}
    }
    \label{fig:Method Overview}
\end{figure*}

\begin{figure}[b]
\vspace{-0.7cm}
    \centering
    \begin{subfigure}[b]{0.3\linewidth}
        \centering
        \includegraphics[width=\linewidth]{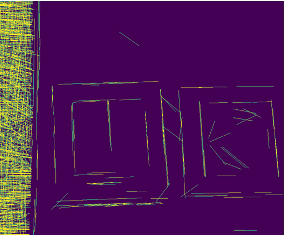}
        \vspace{-0.5cm}
        \caption{}
        \label{fig:1a}
    \end{subfigure}
    \hfill
    \begin{subfigure}[b]{0.3\linewidth}
        \centering
        \includegraphics[width=\linewidth]{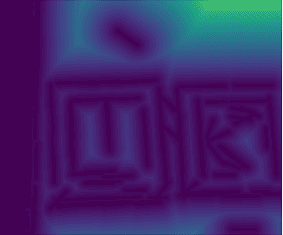}
        \vspace{-0.5cm}
        \caption{}
        \label{fig:1b}
    \end{subfigure}
    \hfill
    \begin{subfigure}[b]{0.3\linewidth}
        \centering
        \includegraphics[width=\linewidth]{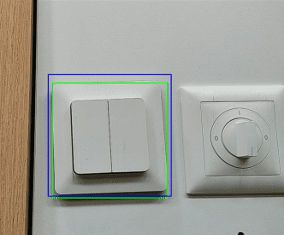}
        \vspace{-0.5cm}
        \caption{}
        \label{fig:1c}
    \end{subfigure}
    \vspace{-0.2cm}
    \caption{\textbf{Distance Transform.} Given the initial image, a line detector is used to detect sharp intensity changes resulting in a). Afterwards a distance transform allows us to generate a smooth map with minima at texture rich regions. Using non-linear, local optimization we optimize the initial bounding box (blue) to better coincide with image edges (green).\vspace{-0.6cm}}
    \label{fig:distance_transform}
\end{figure}

\begin{figure}[t!]
\vspace{-0.4cm}
\centering

 \includegraphics[width=0.95\linewidth]{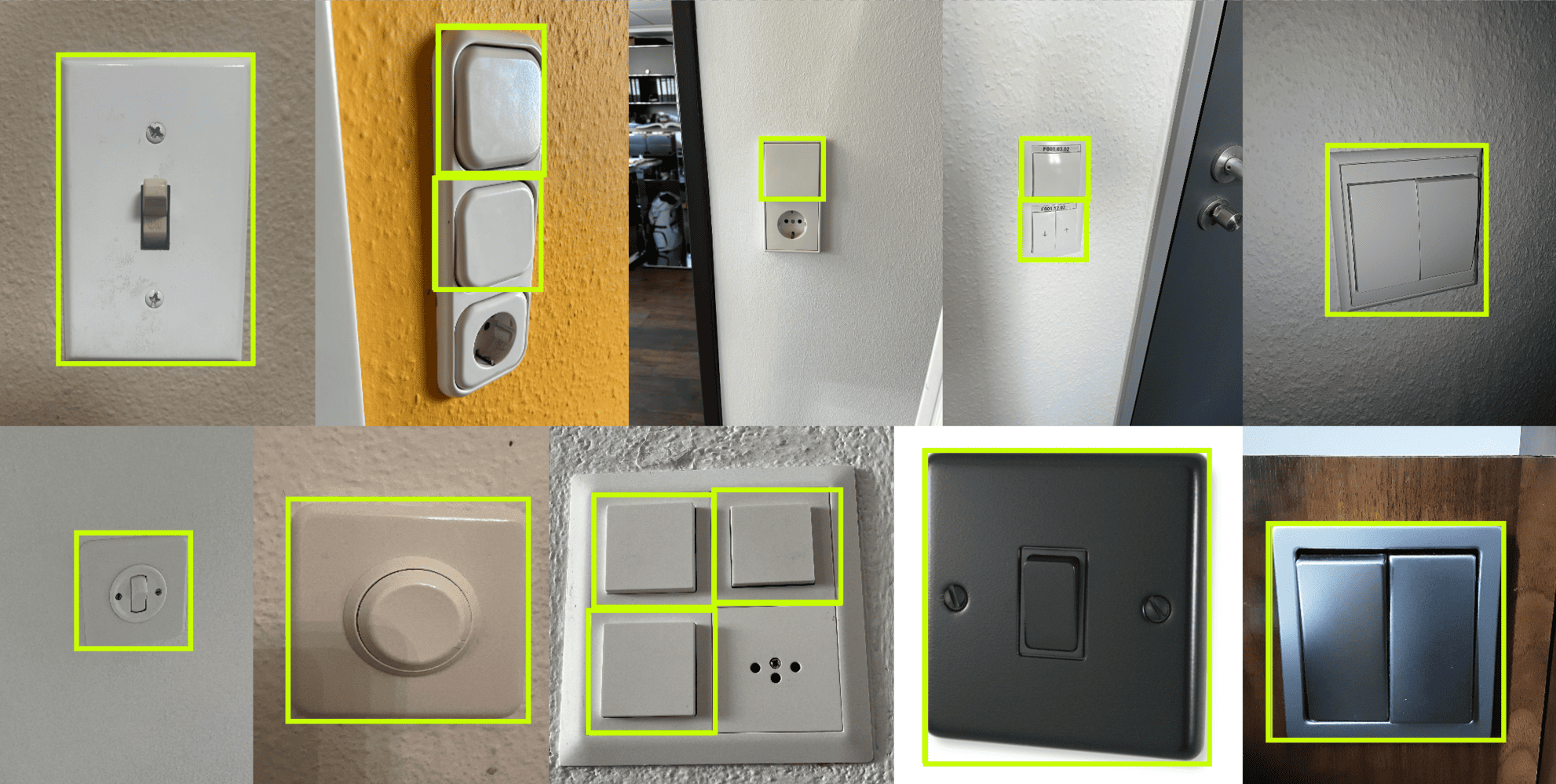}

    \caption{\textbf{Light switch Dataset Overview.} Example of various annotated images in our dataset. It contains a total of 3721 images of light switches, captured at different viewpoints, distances and lighting conditions.  \vspace{-0.7cm}}
    \label{fig:Light Switch Dataset}
\end{figure}

\PAR{Detecting Functional Elements.}
\label{Detecting Functional Elements}
We detect light switches directly from a sequence of images and register these detections with the available 3D point cloud. To do this, we reproject pixel coordinates using the camera transformation provided by the robot’s on-board localization system. We detect switches using a YOLOv8 model \cite{Jocher_Ultralytics_YOLO_2023} which is fine tuned on our custom dataset for light switch detection (See \cref{Dataet and Experiments}). For 3D registration, we assume the 2D projection of the 3D coordinate of the center of the light switch to be located in the geometric center of the bounding box parameterized by $\{x_1,y_1,x_2,y_2\}$ i.e. the upper left and lower right corner of the bounding box. Note that we refer to the center of the light switch's casing, not the center of each individual button (See \cref{fig:Light Switch Dataset} top right, center right) and annotate the data accordingly.

\PAR{Bounding Box Optimization.}
As discussed later (See \cref{Affordance Prediction and Motion Primitives}), our affordance prediction relies on an accurate estimate of the bounding box for handles and switches. 
Although our object detection module provides initial bounding boxes, using them directly often results in failures of the subsequent pose estimation due to inaccuracies of the bounding boxes (i.e. the bounding boxes not being aligned properly with switch lining).
We introduce a bounding box refinement stage as shown in \cref{fig:distance_transform}, aligning them with distinct edge features. 
For each successful detection, we apply a Hough Transform for line detection \cite{duda1972use} to generate a binary image highlighting the line features of the scene. 
By applying a distance transform, we assign every background pixel its euclidian distance from the nearest line pixel. Subsequent normalization of the distance transform yields a 2D distance map $D(x,y) \in [0,1]$ where the minima correspond to the pixel locations of line features. 
The objective of the optimization is to adapt the bounding box such that it coincides with the minima of the distance transform, i.e. edge features. 
Assuming the bounding box to be a good initial guess, we propose a local optimization strategy for the bounding box 
\vspace{-5pt}
\[
\begin{aligned} 
& \underset{z}{\text{arg min}}
& & \mathcal{L}_{dist}(x_1, y_1, x_2, y_2) + \lambda \mathcal{L}_{rect}(x_1, y_1, x_2, y_2) \\ \vspace{-0.4cm}
& \text{s.t.}
& & (x_{1}, y_{1}), (x_{2}, y_{2}) \in \Omega \\
\end{aligned}
\vspace{-0.2cm}
\]
with
\vspace{-0.2cm}
\[
\mathcal{L}_{dist}(x_1, y_1, x_2, y_2) = \frac{1}{N_P} \sum_{(x,y) \in P} D(x,y)
\]
being the mean of the normalized distance transform on the bounding box perimeter $P$, the image pixel describing the bounds of the bounding box(distance loss).
 and

\vspace{-0.3cm}
\[
\mathcal{L}_{\text{rect}}(x_1, y_1, x_2, y_2) = 1 - \frac{\text{A}_{rect}}{\text{A}_{square}} 
\vspace{-0.2cm}
\] 
a cost enforcing squareness , $\lambda \in [0,1]$ a tuning parameter and $\Omega$ the set of all valid image coordinates. $\text{A}_{rect}=(x_2-x_1)(y_2-y_1)$ describes the bounding box area, while $\text{A}_{square}=\max((x_2-x_1),(y_2-y_1))^2$ describes area of the minimum enclosing square.
Note that we introduce this regularization term to increase robustness, based on empirical observations that most light switches are square shaped. 

\PAR{Pose Calculation and Refinement.}
We represent the pose of  a functional element as $\xi=(\vec{r}_c, \vec{n}_c) \in SE(3)$ by its center point $\vec r_{c} \in \mathbb{R}^{3}$ and its interaction normal $\vec n_{c} \in \mathbb{R}^{3}$.
We first estimate an initial pose using an object detector from a far-away image or through a scene graph containing the objects of interest.
This initial "guess" suffers from low positional accuracy since the depth estimation error is greater at far distances. 
Afterwards (as done in \cite{lemke2024spot}) we refine the position closeup. To estimate the center $\vec r_{c}$, we select the geometric center of the bounding box and unproject it onto the point cloud. 
Afterwards, we estimate the interaction normal as the plane normal by reprojecting all corresponding pixels in the bounding box into 3D and using RANSAC for plane estimation \cite{fischler_bolles_1981}. 
We then repeat this process from $N_r$ different close-up viewpoints, estimating $\xi^{i}=(\vec o_{c}^i, \vec n_{c}^i)$ for each viewpoint and compute the final coordinates and plane normal
\vspace{-0.1cm}
\[
\textstyle \xi_{avg} = (\dfrac{1}{N_r} \sum_i^{N_r} \vec r_{c}^{\ i},\ \dfrac{1}{N_r} \sum_i^{N_r} \vec n_{c}^{\ i})
\vspace{-0.1cm}
\] 
as the average over all refinement viewpoints. 
This step improves the robustness significantly (See \cref{tab1:experiments} and \cref{fig:Pipeline Evaluation}), since it averages out estimation errors such as bad bounding boxes, failure to properly estimate interaction normal 
and failure to properly estimate the object center 

\PAR{Affordance Prediction and Motion Primitives.} \label{Affordance Prediction and Motion Primitives}
Having calculated coordinates of the functional centers and plane normal, we predict interaction affordances and calculate the motion primitive necessary for operation. 
In accordance with \cite{delitzas2024scenefun3d, jiang2022, sun2023} we define our motion primitive $\phi(t, \vec a, \vec o)$ where $t=\{\texttt{rotation}, \texttt{translation}\}$ describes the motion type, $\vec{a} \in \mathbb{R}^{3}$ the axis of motion and $\vec{o} \in \mathbb{R}^{3}$ the motion origin. 
Note that the motion origin does not necessarily coincide with the object center $\vec r_{c}$ and that, depending on the type of functional element, multiple motion primitives can be defined for a single object. 
For example, if the object is a light switch with a double push button horizontally aligned. In that case, two distinct primitives can be defined, each with a horizontal offset with respect to the center.
We therefore introduce the set of valid motion primitives as $\Phi = \{\phi_i\}_i^K$, where $K$ denotes the total number of valid interactions.

We leverage a pre-trained VLM (gpt-4-turbo \cite{openai2024gpt4technicalreport}) to predict affordances on a single RGB image of the light switch. These affordances are open-language responses given by the VLM. 
To this end, we prompt the VLM to describe the following characteristics and translate them into 3D Motion Primitives (Quotation marks indicate the VLMs response)\footnote{For more details about the used prompts, please refer to the \href{https://timengelbracht.github.io/SpotLight/}{webpage.}}:
\begin{itemize}[leftmargin=*]
    \item \textbf{Switch Type.} We directly translate the switch type to the motion type: $\texttt{"turn button"} \Rightarrow t=\texttt{rotation}$ and $\{\texttt{"push button", "rocker switch"}\} \Rightarrow t=\texttt{translation}$
    \item \textbf{Button Count and Arrangement.} In case a single switch has multiple buttons, we prompt the VLM to describe button arrangement with respect to each other, which we then translate to offsets $\vec \delta$ along z or y axes $\texttt{"side-by-side"} \Rightarrow \vec o = \vec r_c + \vec \delta_y$ and $\texttt{"stacked vertically"} \Rightarrow \vec o= \vec r_c + \vec \delta_z$ where y and z axes correspond to horizontal and vertical axes respectively. We set the offset to a fixed value, depending on the gripper dimensions. In case the switch only has a single button, we assume it to be located in the center: $\texttt{"single button"} \Rightarrow \vec o_{i}= \vec r_c$.
    \item \textbf{Symbol inference.} Using the VLM's rich visual understanding, we further prompt it to infer interactions from possible symbols present on the light switch. For example, arrow symbols on a push button light switch pointing up and down imply that the switch is a rocker switch affording an eccentric push along the vertical axis: $\texttt{"top/bot push"} \Rightarrow \vec o= r_c + \vec \delta_z$
\end{itemize}
We further assume the axis of motion $\vec{a_{i}} = \vec{n_{c}}$ to coincide with the interaction normal. Note that this does not hold for toggle switches (\cref{Light Switch Dataset}, top left), where the axis of motion lies orthogonal to the interaction normal. While our test setup does not contain such switches, the affordance module can identify them, allowing the integration into the framework.



\vspace{-0.1cm}
\section{Dataset, Experiments and Applications}
\label{Dataet and Experiments}

\PAR{Light Switch Dataset.}\label{Light Switch Dataset}
Even though open-vocabulary detectors such as YOLO-World \cite{cheng2024yolow} exist, we still find their robustness to be lacking when used on object categories outside of the training distribution 
such as light switches (See \cref{tab:Detection Metric}), making fine-tuning on a custom dataset necessary. We gather and annotate a dataset of 715 images of light switches. For this purpose, we mainly collect and annotate the images ourselves, but also incorporate smaller, publicly available light switch datasets such as \cite{light-switch_dataset}. The captured light switches are from regions such as EU, Switzerland and the US. For robustness, images are captured at multiple scales, viewpoints and lighting conditions. All images are resized to 1280x1280 pixels and further augmented. 
After augmentation the dataset contains 3721 images. Note that we used a smaller dataset consisting of 1552 images to train the model YOLOv8$^{1552}$ for all reported experiments in the latter part of the section as the dataset was expanded later. We additionally release the YOLOv8$^{3721}$ model for the benefit of the community. We compare the performance of a YOLO-World model in \cref{tab:Detection Metric} and our fine-tuned model on the test set of our light switch dataset and find a significant increase across all metrics for our fine-tuned model. 

\begin{table}[b]
\vspace{-0.5cm}
\caption{\textbf{\textsc{Light Switch Detection.}} 
We report mAP$_{50}$, mAP$_{50-95}$, precision and recall, comparing the open-vocabularity model YOLO-World to our fine tuned models YOLOv8$^{1552}$ and YOLOv8$^{3721}$. 
}
\vspace{-0.3cm}
\begin{center}
\begin{tabular}{c|c|c|c|c}
\hline
\textbf{Model} & \textbf{\textit{mAP$_{50}$}}& \textbf{\textit{mAP$_{50-95}$}}& \textbf{\textit{precision}} & \textbf{\textit{recall}} \\
\hline
YOLOv8$^{3721}$ &	\fs{0.63} & \fs{0.29} & \fs{0.61} & \fs{0.72} \\
YOLOv8$^{1552}$ &	\nd{0.44} & \nd{0.16} & \nd{0.43} & \nd{0.69} \\
YOLO-World\cite{cheng2024yolow} &	    0.038 & 0.013 & 0.048 & 0.22 \\
\hline
\end{tabular}
\label{tab:Detection Metric}
\end{center}
\end{table}

\PAR{Experimental Setup.}
\begin{figure}[t!]
\vspace{-0.45cm}
    \centering
    \begin{subfigure}[b]{0.57\linewidth}
        \centering
        \includegraphics[width=\linewidth]{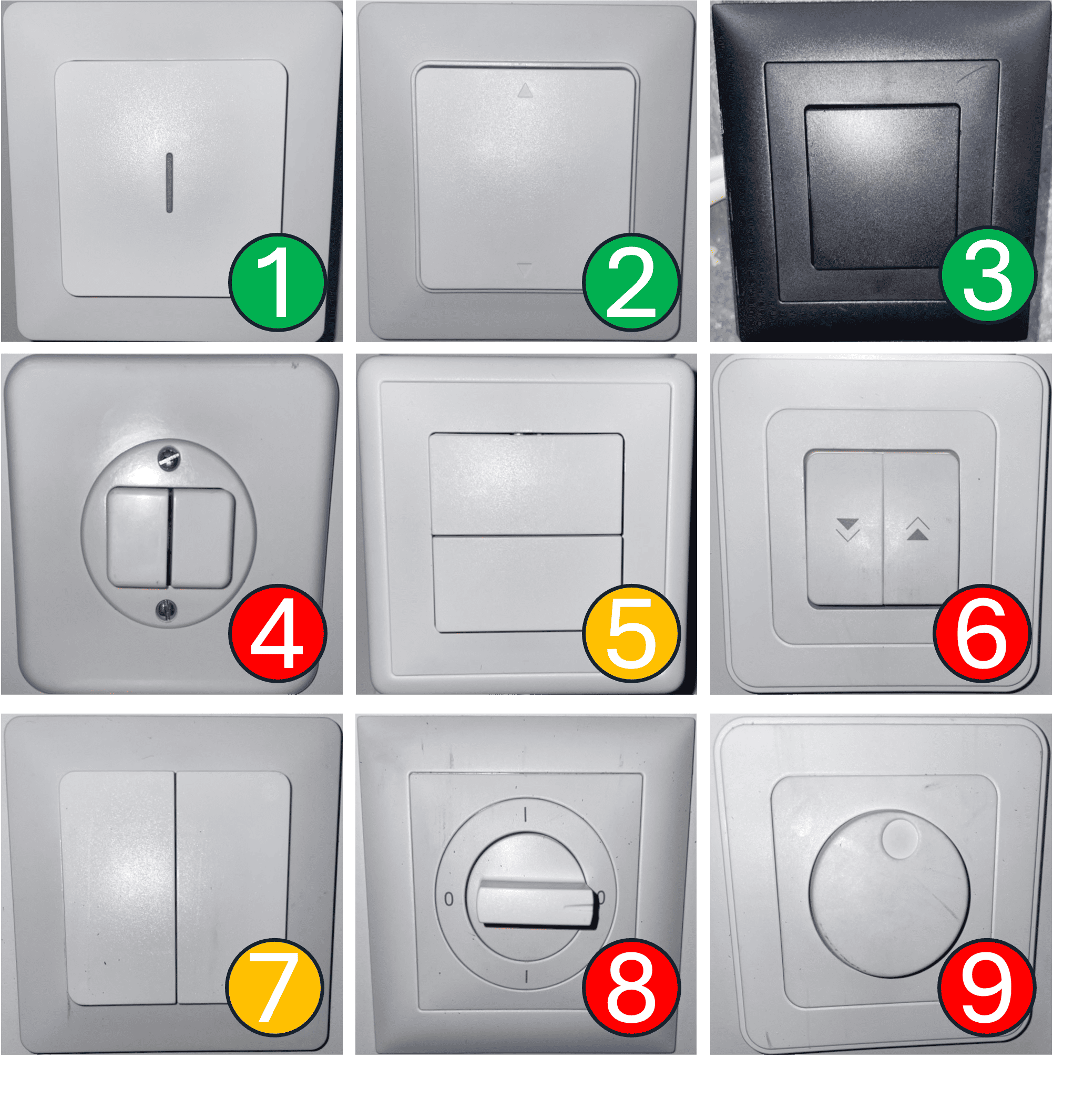}
        \vspace{-0.7cm}
        \caption{}
        \label{fig:Experimental Setup a}
    \end{subfigure}
    \hfill
    \begin{subfigure}[b]{0.41\linewidth}
        \centering
        \includegraphics[width=\linewidth]{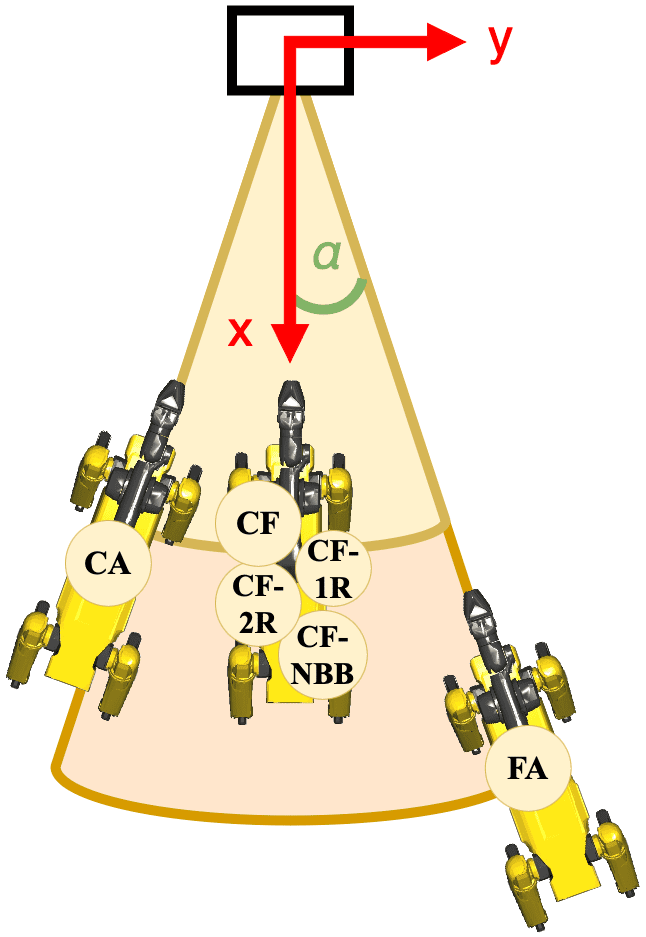}
        \vspace{-0.7cm}
        \caption{}
        \label{fig:Experimental Setup b}
    \end{subfigure}
    \vspace{-0.6cm}
    \caption{\textbf{Experimental Setup.} \textbf{a)} Light switches used for framework evaluation. The number shows the index for Per-switch Evaluations, with difficulty levels color-coded (green - easy, yellow - medium, red - hard). \textbf{b)} Test rig (black box) and robot positioning during experiments, indicated by abbreviations on the robot’s back.\vspace{-0.8cm}
    } 
    \label{fig:Experimental Setup}
\end{figure}
To test our method in a real world scenario we install nine of the most common light switches that are permitted to install in Switzerland on an IKEA KALLAX shelving system. For evaluation of the generalizability of our method we selected light switches of varying form, interaction type, and colors, as shown in \cref{fig:Experimental Setup a}.  
Further, to facilitate gripping of small protruding parts, as seen in turning switches, we 3D print a removable (press-fit) gripper extension to the robots gripper. 
We devise two series of experiments to evaluate overall framework performance as well as the impact of its subsystems. To this end we introduce a polar coordinate system, as seen in \cref{fig:Experimental Setup}, with the testrig at the origin, defining the robot's position $p_{i}(r, \alpha)$ by its radial distance $r$ and polar angle $\alpha$. The success rate $SR = N_{success}/N_{attempt}$ is defined as the number of successful light switch interactions over the number of attempts. We classify an attempt only as successful if the light switch is detected, the interaction affordance correctly predicted and the switch successfully operated. For each experiment, every light switch in the testrig is attempted 10 times for a total of $N_{attempt}=90$ attempts per experiment.

\PAR{Ablation Study.}In a first series of experiments the overall performance of the entire framework is evaluated.  Firstly, we assess the overall success rate close to the ligth-switches without angular disturbance (\textbf{C}lose Distance - \textbf{F}ixed position - \textbf{CF}). Secondly, we assess the influence of angular disturbance by varying the robots heading from $-22.5^\circ$ to $22.5^\circ$. 
(\textbf{C}lose Distance - \textbf{A}ngular position \textbf{CA}). 
Lastly, we investigate the robustness of the framework under angular disturbances as well as greater distance by increasing the robots radial distance to $2m$ (\textbf{F}ar Distance - \textbf{A}ngular position - \textbf{FA}).


\begin{table}[b]
 \vspace{-0.6cm}
\caption{\textbf{\textsc{Experiments}} We evaluate the Full Framework under varying initial distances and angles by measuring the success rate for button interaction. Further experiments evaluate the influence of the number of pose refinements and the bounding box refinement on the overall success rate. CI$_{95}$ indicates the $95\%$ confidence interval.}
\vspace{-0.3cm}
\begin{center}
\begin{tabular}{c|c|c|c|c|c}
		\toprule
\multicolumn{6}{c}{\textbf{Full Framework}} \\ \hline
\textbf{Exp.} & \textbf{\textit{r $[m]$}}& \textbf{\textit{$\alpha$ $[^\circ]$}}& PR & BB &\textbf{\textit{SR (CI$_{95}$)}} \\
\toprule
\textbf{CF} & 1.5& 0& 4 & \cmark & \fs{84\% (92\%,	77\%)} \\
\textbf{CA} & 1.5& $\pm 22.5^{\circ}$& 4 & \cmark & \nd{79\% (87\%, 71\%)} \\
\textbf{FA} & 2.0& $\pm22.5^{\circ}$& 4 & \cmark & 68\% (78\%, 59\%) \\
\hline
\multicolumn{6}{c}{\textbf{Partial Framework}} \\
\hline
\textbf{CF-NBB} &  1.5& 0& 4& \xmark & \fs{77\% (85\%, 68\%)} \\
\textbf{CF-2R} &  1.5& 0& 2& \cmark & \nd{67\% (76\%, 57\%)} \\
\textbf{CF-1R} &  1.5& 0& 1& \cmark & 52\% (63\%, 42\%) \\
 \bottomrule
\end{tabular}
\label{tab1:experiments}
\end{center}
\vspace{-0.5cm}
\end{table}


\begin{figure}[t!]
\vspace*{-0.3cm}
	\centering
	\def\svgwidth{\linewidth} 
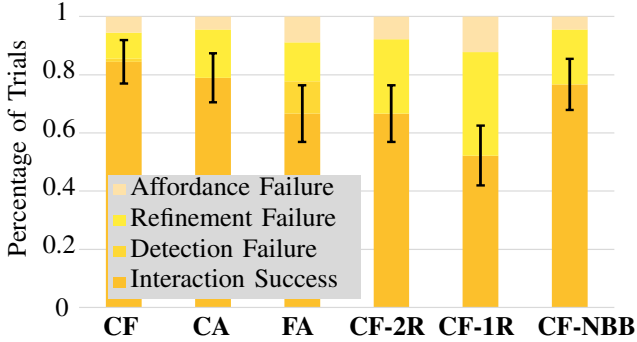
\vspace*{-0.8cm}
        \caption{\textbf{Pipeline Evaluation.} We classify all either as successful or failed, while differentiating between Detection Failure, Refinement Failure and Affordance Failure. One can observe that angular disturbances (\textbf{CA, FA}) a lower number of refinement poses (\textbf{CF-2R, CF-1R}) and no bounding box optimization (\textbf{CF-NBB}) all lead to increased refinement failure. Increases in detection distance mainly lead to increases in detection failures (\textbf{FA}), while affordance failure rates stay comparatively constant through all experiments.
        }
	    \vspace{-0.6cm}	\label{fig:Pipeline Evaluation}
\end{figure}

Failure behaviour can be split into three parts (See \cref{fig:Pipeline Evaluation}): Detection failure, where a light switch is present but not detected; refinement failure, involving incorrect or missing 3D pose estimation; and affordance failure, where the module misclassifies the switch type, leading to incorrect interaction.
In the \textbf{CF} experiment, we position the robot at a fixed distance in front of the test rig, and measure a success rate of $84\%$ over the entire process spanning initial detection and registration, close-up pose refinement, affordance prediction and, lastly, successful interaction. A large share of the failed interactions are caused by either refinement or affordance failure, as displayed in \cref{fig:Pipeline Evaluation}. 
We achieve a lower success rate of $79\%$ if we position the robot at an angle in Experiment \textbf{CA}, which is mainly due to a higher rate of refinement failure. The success rate further drops to $68\%$ if the initial detection distance is increased (Experiment \textbf{FA}) due to erroneous detections at far away distances. Especially the black switch (See \cref{fig:Experimental Setup}) cannot be accurately detected for a lack of visible texture.
A second series of experiments is used to evaluate the impact of pose refinement (PR) and bounding box optimization on overall system performance. The robot position is kept fixed at $p_{4,5,6}(r=1.5m, \alpha=0^{\circ})$ for all experiments of this series. In the first two experiments we investigate the effect of the pose refinements by reducing the number of pose refinements from $4$ to $2$ and finally to $1$ (\textbf{C}lose Distance - \textbf{F}ixed position - \textbf{R}educed number of  pose refinements \textbf{CF-xR}). In the last experiment, we run the framework without optimizing the bounding boxes(\textbf{C}lose Distance - \textbf{F}ixed position- \textbf{N}o \textbf{B}ounding \textbf{B}ox Optimization - \textbf{CF-NBB}).

The success rate drops significantly to $67\%$ and $52\%$ if we lower the number of refinements to $2$ or $1$ pose(s). mainly due to an increase in refinement failures, particularly in estimating the plane normal. Especially the rotating switches, whose surface deviates the most from the flat plane surface, contribute significantly to this decrease in success rate. Lastly, one can observe a decrease of success rate to $77\%$ if we omit the bounding box optimization from the framework (\textbf{CF-NBB}). This can also be traced back to an increase in refinement failures, due to the light switch center being estimated incorrectly. 

\begin{figure*}
\vspace{-0.6cm}
    \centering
    \includegraphics[width=0.9\linewidth]{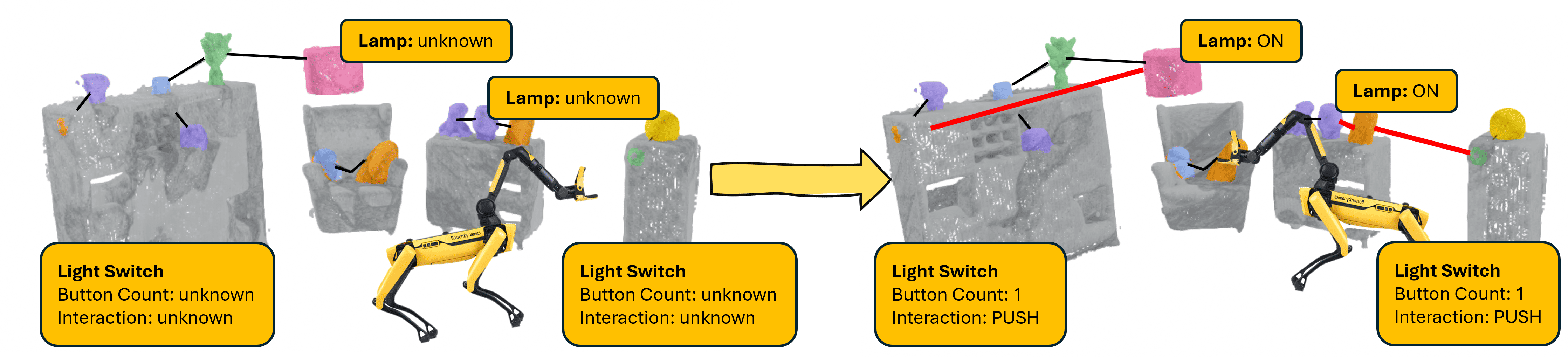}
    \vspace{-0.2cm}
    \caption{\textbf{Learning through Interactions.} Using the SpotLight Framework, the robot refines its knowledge about the light switches: It refines the switch pose and offers a more detailed description of the switch type and what interation it affords. Also, it uncovers underlying relationships between lamps and switches (red edges) by interacting with switches and monitoring changes in the lamps' intensities, thus offering a richer scene representation.\vspace{-0.6cm}
    }
    \label{fig:Updated Scene Graph}
\end{figure*}

\PAR{Learning Scene Graphs through Interaction.}
Having demonstrated the applicability of our framework for detecting and interacting with light switches in a lab setup, we further show that our framework enables the robot to expand its own understanding of the scene through interaction. The main goal of this application is to demonstrate the robot's capability, facilitated by our framework, to learn the relationship between light switches and lamps present in the scene, given a scene graph containing only the object's spatial information. 

\textit{Building the Scene Graph.}
To this end we introduce a scene graph $G=(V, E)$, a lightweight representation of the scene, with $V=\{v_{1},...,v_{n} | v_i = (\xi_i, c_i, \Phi_i, s_i)\}$ being the set of vertices representing the scene's objects and set of edges $E$, representing relationships between objects. Each vertex is described by a pose $\xi_{i}=(\vec o_{c}, \vec n_{c})$, object class $c_i \in \{1,...,N_c\}$, a set of motion primitives $\Phi_i =\{\phi_{j}\}_{j=1}^{K}$ and an optional state $ s_i \in \{\texttt{on}, \texttt{off}\, \texttt{none}\} $. The given scene graph is built by first capturing point cloud of the scene, while also capturing RGB frames and Poses. We then segment the scene using Openmask3D \cite{takmaz2023openmask3d}. Due to the lower spacial resolution of the point clouds compared to the images, small functional elements such as light switches are missed in the 3D segmentation. Therefore, to register the light switches in the scene graph, we run our light switch detector (See \cref{Detecting Functional Elements}) on the RGB frames and lift these to 3D. To handle multiple detections of the same light switch across frames, we reproject each detection into 3D and cluster them so each group represents a single switch. For each cluster, the detection with the highest confidence and its 3D coordinates are then registered into the scene graph.

\textit{Extending the Scene Graph.}
We extend the scene graph by having the robot explore the relationships between light switches and lamps, much like a human would. Given no information except for spatial arrangement, the robot autonomously operates a light switch, while checking the state of each lamp in the scene before and after operation.
For simplicity we omit dimmer switches and therefore assume a binary lamp state $s_{l_i} \in \{\texttt{on}, \texttt{off}\}$. To infer the state after light switch operation query the VLM with two images of a corresponding lamp, one before and one after interaction and output the state change (\texttt{off} to \texttt{on}, \texttt{on} to \texttt{off}, \texttt{no change})\footnote{Using the before and after image to infer a state change is necessary because the VLM struggles to infer an absolute state from a single image.}.
Given the robot operates a switch $v_i = (\xi_i, \texttt{switch}, \Phi_i, s_i)$ and detects a state change in a specific lamp $v_j = (\xi_j, \texttt{lamp}, \Phi_j, s_j)$ we can adapt the scene graph accordingly: Firstly, we register the newly found relationship by introducing a new edge $(v_i, v_j)$ and adapting the edge space $E' \leftarrow (v_i, v_j) \cup E$ (See \cref{fig:Updated Scene Graph}). Secondly, we assign the lamp its new state: $s_j \leftarrow s_j'$ . Furthermore, we can also refine switch information $\Phi_i  \leftarrow \{\phi_{i}\}_{i=1}^{K}$ using SpotLight's affordance detection. This formulation can even capture wirings where multiple switches control the same lamp, simply with multiple edges per lamp. 

\PAR{Extension to Swing Doors.} We further introduce an extension to swing doors and drawers, showing the versatility of our approach. 
Similarly as done in \cite{lemke2024spot,jiang2022}, we describe the interaction by a set of two motion primitives: 
Firstly, a linear motion along an axis (prismatic motion), as presented in drawer interactions or push button switches and secondly a rotating motion around an axis (revolute motion).

Using the same 2D approach for detection, we detect drawer and handles independently using the fine-tuned YOLOv8 introduced in the spot-compose framework \cite{lemke2024spot} and thereafter reprojecting pixel coordinates onto the given point cloud and rely on the postprocesing from \cite{lemke2024spot}. Depending on the type and orientation of the handle, the gripper roll needs to be adapted such that it aligns with the major axis of the handle. We compute the major handle axis using principal component analysis (PCA) \cite{Pearson_1901} on the handle point cloud.

\textit{Affordance Detection and Motion Primitive.} Since the interaction with drawers and swing doors is more straightforward compared to light switch interaction, we predict the affordance solely based on the geometric relationship of handle and drawer/door, thus neglecting VLM-based affordance prediction. Note that, one could still use the VLM for affordance prediction by adapting the query and output format accordingly. Given the bounding boxes of both handle and drawer we build the motion primitive necessary for interaction as follows: We assume motion type $t=\texttt{rotation}$ if the handle is positioned eccentrically within the door bounding box. 
We calculate the axis of motion as the axis aligned with the far edge of the bounding box (hinge axis). Further, we define the motion origin $\vec{o}= \vec r_c$ as the handle center. 
Contrary to light switches, we also define a metric lever $l$ as the orthogonal distance of the handle to the axis of motion as well as a direction of rotation. 
Finally, we track the circular motion trajectory parameterized by the door opening angle $\alpha \in [0, \frac{\pi}{2}]$ with $\prescript{}{G}{\tau (\alpha)} = -l \cdot [\ \sin(\alpha), \pm (1 - \cos(\alpha)), 0\ ] $ in the initial handle frame $G$. Considering the scene graph representation presented earlier, one can introduce swing door nodes $v_j = (\xi_j, \texttt{swing door}, \Phi_j, s_j)$ with an adapted state $s_j = \alpha$ describing the current door angle. We refer to the \href{https://timengelbracht.github.io/SpotLight/}{project webpage} for further information.



\section{Conclusion and Limitations}
We present \emph{SpotLight}, a framework for interacting with light-switches and other functional elements using a mobile agent. 
Our proposed method introduces a new Light-Switch dataset and fine-tuned YOLOv8 detector for light-switch detection. 
By combining VLM-based affordance predictions together with 3D pose refinements we are able to estimate motion primitives for interaction.
We extensively evaluate our approach on a large variety of light switches and further, demonstrate the utility of our framework for embodied learning through interaction.
By interacting with light switches and inspecting lamp states, we demonstrate the agent’s ability to learn hidden relationships within a scene graph using interactive perception. Additionally, we show how our framework can open swing doors and interact with articulated objects in domestic settings via the Spot-Compose framework \cite{lemke2024spot}.  While real-world tests with a mobile robot show promising results, the open-loop approach using motion primitives is sensitive to pose estimation failures
Lastly, we would like to emphasize the  VLM's affordance prediction capabilities, showing strong zero-shot capabilities and low failure rates. We note that key to these success rates is to prompt the VLMs with good close-up images of the functional element in question, emphasizing a need for good detectors. We aim to further investigate using VLM's for affordance in more complex setups.


\bibliography{references}

\end{document}

%% file: data/img/barchart_experiments_1.pdf_tex
\begingroup%
  \makeatletter%
  \providecommand\color[2][]{%
    \errmessage{(Inkscape) Color is used for the text in Inkscape, but the package 'color.sty' is not loaded}%
    \renewcommand\color[2][]{}%
  }%
  \providecommand\transparent[1]{%
    \errmessage{(Inkscape) Transparency is used (non-zero) for the text in Inkscape, but the package 'transparent.sty' is not loaded}%
    \renewcommand\transparent[1]{}%
  }%
  \providecommand\rotatebox[2]{#2}%
  \newcommand*\fsize{\dimexpr\f@size pt\relax}%
  \newcommand*\lineheight[1]{\fontsize{\fsize}{#1\fsize}\selectfont}%
  \ifx\svgwidth\undefined%
    \setlength{\unitlength}{1100.84976196bp}%
    \ifx\svgscale\undefined%
      \relax%
    \else%
      \setlength{\unitlength}{\unitlength * \real{\svgscale}}%
    \fi%
  \else%
    \setlength{\unitlength}{\svgwidth}%
  \fi%
  \global\let\svgwidth\undefined%
  \global\let\svgscale\undefined%
  \makeatother%
  \begin{picture}(1,0.54498942)%
    \lineheight{1}%
    \setlength\tabcolsep{0pt}%
    \put(0,0){\includegraphics[width=\unitlength,page=1]{barchart_experiments_1.pdf}}%
    \put(0.08190564,0.07061798){\color[rgb]{0,0,0}\makebox(0,0)[lt]{\lineheight{1.25}\smash{\begin{tabular}[t]{l}0\end{tabular}}}}%
    \put(0.06281865,0.16556928){\color[rgb]{0,0,0}\makebox(0,0)[lt]{\lineheight{1.25}\smash{\begin{tabular}[t]{l}0.2\end{tabular}}}}%
    \put(0.06281865,0.25331616){\color[rgb]{0,0,0}\makebox(0,0)[lt]{\lineheight{1.25}\smash{\begin{tabular}[t]{l}0.4\end{tabular}}}}%
    \put(0.06281865,0.34054073){\color[rgb]{0,0,0}\makebox(0,0)[lt]{\lineheight{1.25}\smash{\begin{tabular}[t]{l}0.6\end{tabular}}}}%
    \put(0.06281865,0.4277653){\color[rgb]{0,0,0}\makebox(0,0)[lt]{\lineheight{1.25}\smash{\begin{tabular}[t]{l}0.8\end{tabular}}}}%
    \put(0.08190564,0.51551217){\color[rgb]{0,0,0}\makebox(0,0)[lt]{\lineheight{1.25}\smash{\begin{tabular}[t]{l}1\end{tabular}}}}%
    \put(0.15363064,0.04241362){\color[rgb]{0,0,0}\makebox(0,0)[lt]{\lineheight{1.25}\smash{\begin{tabular}[t]{l}\textbf{CF}\end{tabular}}}}%
    \put(0.28751115,0.04241362){\color[rgb]{0,0,0}\makebox(0,0)[lt]{\lineheight{1.25}\smash{\begin{tabular}[t]{l}\textbf{CA}\end{tabular}}}}%
    \put(0.42139166,0.04241362){\color[rgb]{0,0,0}\makebox(0,0)[lt]{\lineheight{1.25}\smash{\begin{tabular}[t]{l}\textbf{FA}\end{tabular}}}}%
    \put(0.52257012,0.04241362){\color[rgb]{0,0,0}\makebox(0,0)[lt]{\lineheight{1.25}\smash{\begin{tabular}[t]{l}\textbf{CF}\textbf{-2R}\end{tabular}}}}%
    \put(0.65645042,0.04241331){\color[rgb]{0,0,0}\makebox(0,0)[lt]{\lineheight{1.25}\smash{\begin{tabular}[t]{l}\textbf{CF}\textbf{-1R}\end{tabular}}}}%
    \put(0.80531961,0.04241362){\color[rgb]{0,0,0}\makebox(0,0)[lt]{\lineheight{1.25}\smash{\begin{tabular}[t]{l}\textbf{CF}\textbf{-NBB}\end{tabular}}}}%
    \put(0.03739719,0.15940564){\color[rgb]{0,0,0.00392157}\rotatebox{90}{\makebox(0,0)[lt]{\lineheight{1.25}\smash{\begin{tabular}[t]{l}Percentage of Trials\end{tabular}}}}}%
    \put(0,0){\includegraphics[width=\unitlength,page=2]{barchart_experiments_1.pdf}}%
    \put(0.19472527,0.25279385){\color[rgb]{0,0,0}\makebox(0,0)[lt]{\lineheight{1.25}\smash{\begin{tabular}[t]{l}Affordance Failure\end{tabular}}}}%
    \put(0,0){\includegraphics[width=\unitlength,page=3]{barchart_experiments_1.pdf}}%
    \put(0.19472527,0.2063089){\color[rgb]{0,0,0}\makebox(0,0)[lt]{\lineheight{1.25}\smash{\begin{tabular}[t]{l}Refinement Failure\end{tabular}}}}%
    \put(0,0){\includegraphics[width=\unitlength,page=4]{barchart_experiments_1.pdf}}%
    \put(0.19472527,0.16034626){\color[rgb]{0,0,0}\makebox(0,0)[lt]{\lineheight{1.25}\smash{\begin{tabular}[t]{l}Detection Failure\end{tabular}}}}%
    \put(0,0){\includegraphics[width=\unitlength,page=5]{barchart_experiments_1.pdf}}%
    \put(0.19472527,0.11386131){\color[rgb]{0,0,0}\makebox(0,0)[lt]{\lineheight{1.25}\smash{\begin{tabular}[t]{l}Interaction Success\end{tabular}}}}%
  \end{picture}%
\endgroup%